\documentclass[11pt,a4paper]{article}
\usepackage[hyperref]{emnlp2020}
\usepackage{times}
\usepackage{latexsym}
\usepackage{url}
\usepackage{xcolor}
\usepackage{amssymb}
\usepackage{amsmath}
\usepackage{array}
\usepackage{booktabs}
\usepackage{graphicx}
\usepackage{enumitem}
\usepackage{colortbl}
\usepackage{svg}
\usepackage{multirow}
\usepackage{bm}
\usepackage{hyperref}
\usepackage{array}

\newcolumntype{P}[1]{>{\centering\arraybackslash}p{#1}}

\definecolor{darkspringgreen}{rgb}{0.09, 0.45, 0.27}
\usepackage{microtype}

\aclfinalcopy

\title{Paraphrasing vs Coreferring: Two Sides of the Same Coin}

\author{Yehudit Meged$^1$~~~Avi Caciularu$^{1,2}$~~~Vered Shwartz$^{2,3}$~~~Ido Dagan$^1$ \\
    $^1$Computer Science Department, Bar-Ilan University, Ramat-Gan, Israel\\
    $^2$Allen Institute for Artificial Intelligence\\
    $^3$Paul G. Allen School of Computer Science \& Engineering, University of Washington\\       
    {\tt\small \{yehuditmeged,avi.c33\}@gmail.com, vereds@allenai.org, dagan@cs.biu.ac.il} \\
}

\date{}

\begin{document}
\maketitle

\begin{abstract}
We study the potential synergy between two different NLP tasks, both confronting predicate lexical variability: identifying predicate paraphrases, and event coreference resolution. First, we used annotations from an event coreference dataset as distant supervision to re-score heuristically-extracted predicate paraphrases. The new scoring gained more than 18 points in average precision upon their ranking by the original scoring method. Then, we used the same re-ranking features as additional inputs to a state-of-the-art event coreference resolution model, which yielded modest but consistent improvements to the model's performance. The results suggest a promising direction to leverage data and models for each of the tasks to the benefit of the other.
\end{abstract}

\section{Introduction}
\label{sec:intro}
Recognizing that mentions of different lexical predicates discuss the same event is challenging \cite{barhom-etal-2019-revisiting}. Lexical resources such as WordNet \cite{miller1995wordnet} capture such synonyms (\textit{say, tell}) and hypernyms (\textit{whisper, talk}), as well as antonyms, which can be used to refer to the same event when the arguments are reversed ({[}a]$_0$ beat {[}a]$_1$, {[}a]$_1$ lose to {[}a]$_0$). However, WordNet's coverage is insufficient, in particular, missing context-specific paraphrases (e.g. (\textit{hide, launder}), in the context of money).  Conversely, distributional methods enjoy broader coverage, but their precision for this purpose is limited because distributionally similar terms may often be mutually-exclusive (\textit{born, die}) or may refer to different event types which are only temporally or causally related (\textit{sentenced, convicted}).

Two prominent lines of work pertaining to identifying predicates whose meaning or referents can be matched are cross-document (CD) event coreference resolution and recognizing predicate paraphrases. The former identifies and clusters event mentions, across multiple documents, that refer to the same event \emph{within their respective contexts}. The latter task, on the other hand, collects pairs of event expressions that, at the generic lexical level, may refer to the same event \emph{in certain contexts}. Table~\ref{table:ecb_vs_chirps} illustrates this difference with examples of co-referable predicate paraphrases, while their mentions obviously do not always co-refer. 

\setlength\tabcolsep{1pt}
\begin{table}[!t]
    \centering
    \small
        \begin{tabular}{ll}
        \toprule
        Tara Reid has \underline{checked into}\textcolor{darkspringgreen}{$_{\bm{\vee}}$} Promises Treatment Center.\\ 
        Actress Tara Reid \underline{entered}\textcolor{darkspringgreen}{$_{\bm{\vee}}$} well-known Malibu rehab center.\\
        Lindsay Lohan \underline{checked into}\textcolor{red}{$_{\bm{\times}}$} rehab in Malibu, California.\\
        \midrule
        Director Chris Weitz is expected to \underline{direct}\textcolor{darkspringgreen}{$_{\bm{\vee}}$} New Moon. \\ 
        Chris Weitz will \underline{take on}\textcolor{darkspringgreen}{$_{\bm{\vee}}$} the sequel to ``Twilight''.\\ 
        Gary Ross is still in negotiations to \underline{direct}\textcolor{red}{$_{\bm{\times}}$} the sequel.\\
        \bottomrule
        \end{tabular}
    \caption{Examples from ECB+ (a cross-document coreference dataset) that illustrate the context-sensitive nature of event coreference. The illustrated predicates are co-referable, and hence may be used to refer to the same event in certain contexts, but obviously not all their mentions corefer.}
    \label{table:ecb_vs_chirps}
\end{table}

Cross-document event coreference resolution systems are typically supervised, usually trained on the ECB+ dataset, which contains clusters of news articles on different topics \cite{Cybulska2014UsingAS}. Recent systems rely on neural representations of the mentions and their contexts \cite{kenyon-dean-etal-2018-resolving,barhom-etal-2019-revisiting}, while earlier approaches leveraged WordNet and other lexical resources to obtain a signal of whether a pair of mentions \textit{may} be coreferring \cite[e.g.][]{bejan-harabagiu-2010-unsupervised,yang2015hierarchical}. 

Approaches for acquiring predicate paraphrase, in the form of a pair of paraphrastic predicates or predicate templates, were based mostly on unsupervised signals. These included similarity between argument distributions \cite{Lin2001DIRTD, berant2012global}, backtranslation across languages \cite{barzilay-mckeown-2001-extracting,ganitkevitch2013ppdb,mallinson-etal-2017-paraphrasing}, or leveraging redundant news reports on the same event, which are hence likely to refer to the same events and entities using different words \cite{shinyama2002automatic,shinyama-sekine-2006-preemptive,barzilay-lee-2003-learning,zhang2013harvesting,xu2014extracting,shwartz-etal-2017-acquiring}. In some cases, the paraphrase collection phase includes a step of validating a subset of the paraphrases and training a model on these gold paraphrases to re-rank the entire resource \cite{lan-etal-2017-continuously}.     

In this paper, we study the potential synergy between predicate paraphrases and event coreference resolution. We show that the data and models for
one task can benefit the other. In one direction (Section \ref{sec:improving_chirps}), we use event coreference annotations from the ECB+ dataset as distant supervision to learn an improved scoring of predicate paraphrases in the unsupervised Chirps resource \cite{shwartz-etal-2017-acquiring}. The distantly supervised scorer significantly improves upon ranking by the original Chirps scores, adding
18 points to average precision over a test sample.

In the other direction (Section \ref{sec:using_chirps}), we incorporate data from Chirps, represented in the Chirps re-scorer feature vector, into a state-of-the-art event coreference system \cite{barhom-etal-2019-revisiting}. Chirps has a substantial coverage over the ECB+ coreferring mention pairs, and consequently, the incorporation 
yields a modest but consistent improvement across the various coreference metrics.\footnote{Code available at \href{https://github.com/yehudit96/coreferrability}{github.com/yehudit96/coreferrability},
\href{https://github.com/yehudit96/event_entity_coref_ecb_plus}{github.com/yehudit96/event\_entity\_coref\_ecb\_plus}}

\section{Background and Motivation}
\label{sec:related_work}
In this section we provide some background about the cross-document coreference resolution and paraphrase identification (acquisition) tasks, which is relevant for our approaches for synergizing these two tasks. 

\subsection{Event Coreference Resolution}
\label{sec:bg_event_coref}

Event coreference resolution aims to identify and cluster event mentions, that, within their respective contexts, refer to the same event. The task has two variants, one in which coreferring mentions are within the same document (within document) and another in which corefering mentions may be in different documents (cross-document, CD), on which we focus in this paper. 

The standard datasets used for CD event coreference training and evaluation are ECB+ \cite{Cybulska2014UsingAS}, and its predecessors, EECB \cite{lee-etal-2012-joint} and ECB \cite{bejan-harabagiu-2010-unsupervised}. ECB+ contains a set of topics, each containing a set of documents describing the same global event. Both event and entity coreferences are annotated in ECB+, within and across documents.

Models for CD event coreference utilize a range of features, including lexical overlap among mention pairs and semantic knowledge from WordNet \cite{bejan-harabagiu-2010-unsupervised,bejan2014unsupervised,yang2015hierarchical}, distributional \cite{choubey-huang-2017-event} and contextual representations \cite{kenyon-dean-etal-2018-resolving,barhom-etal-2019-revisiting}.

The current state-of-the-art model from \newcite{barhom-etal-2019-revisiting} iteratively and intermittently learns to cluster events and entities. A mention representation $m_i$ consists of several components, representing both the mention span and its surrounding context. The interdependence between clustering event vs. entity mentions is encoded into the mention representation, such that an event mention representation contains a component reflecting the current entity clustering, and vice versa. Using this representation, the model trains a pairwise mention scoring function that predicts the probability that two mentions refer to the same event. 

\subsection{Paraphrase Identification and acquisition}
\label{sec:bg_identification}

Paraphrases are differing textual realizations of the same meaning \cite{ganitkevitch2013ppdb}, typically phrases or sentences \cite{dolan2005microsoft}. A prominent approach for identifying and collecting paraphrases, \emph{backtranslation}, assumes that if two (say) English phrases translate to the same term in a foreign language, across multiple foreign languages, this indicates that these two phrases are paraphrases. This approach was first suggested by \newcite{barzilay-mckeown-2001-extracting}, later adapted to acquire the large PPDB resource \cite{ganitkevitch2013ppdb}, and was also shown to work well with neural machine translation \cite{mallinson-etal-2017-paraphrasing}. 

\paragraph{Paraphrase Identification through Event Coreference.} An alternative approach for paraphrase identification, on which we focus in this paper, leverages multiple news documents discussing the same event. The underlying assumption is that such redundant texts may refer to the same entities or events using lexically-divergent mentions. Co-referring mentions are identified heuristically and extracted as candidate paraphrases. When long documents are used, the first step in this approach is to align each pair of documents by sentences. This was done by finding sentences with shared named entities \cite{shinyama2002automatic} or lexical overlap
\cite{barzilay-lee-2003-learning,shinyama-sekine-2006-preemptive},  and by aligning pairs of predicates or arguments
\cite{zhang2013harvesting,recasens-etal-2013-referent}. In more recent work, \newcite{xu2014extracting} and \newcite{lan-etal-2017-continuously} extracted sentential paraphrases from Twitter by heuristically matching pairs of tweets discussing the same topic.

\paragraph{Predicate Paraphrases.} In contrast to sentential paraphrases, it is also beneficial to identify differing textual \emph{templates} of the same meaning. In this paper we focus on binary predicate paraphrases such as (``[a$_0$] quit from [a$_1$]'', ``[a$_0$] resign from [a$_1$]''). 

Earlier approaches for acquiring predicate paraphrases considered a pair of predicate templates as paraphrases if the distributions of their argument instantiations were similar. For instance, in ``[a$_0$] quit from [a$_1$]'', [a$_0$] would typically be instantiated by people names while [a$_1$] by employer organizations or job titles. A paraphrastic template like ``[a$_0$] resign from [a$_1$]'' is hence expected to have similar argument distributions, and can thus be detected by a distributional similarity approach \cite{Lin2001DIRTD,szpektor2004scaling,berant2012global}. Yet, as mentioned earlier, predicates with similar argument distributions are not necessarily paraphrastic, which introduces a substantial level of noise when acquiring paraphrase pairs using this approach. 

In this paper, we follow the potentially more reliable paraphrase acquisition approach, which tries to heuristically identify concrete co-referring predicate mentions. Identifying such mention pairs,  detected as actually being used to refer to the same event, can provide a strong signal for identifying these predicates as paraphrastic (vs. the quite noisy corpus-level signal of distributional similarity). In particular, we utilize the Chirps paraphrase acquisition method and resource, which follows this approach as described next in some detail.

\paragraph{Chirps: a Coreference-Driven Paraphrase Resource.} Chirps \cite{shwartz-etal-2017-acquiring} is a resource of predicate paraphrases extracted heuristically from Twitter. Chirps aims to recognize coreferring events by relying on the redundancy of news headlines posted on Twitter on the same day. It extracts binary predicate-argument tuples from each tweet and aligns pairs of predicate mentions whose arguments match, by some lexical matching criteria. The matched pairs of arguments are termed \emph{supporting pairs}, e.g. (\textit{Chuck Berry}, \textit{90}) for ``[Chuck Berry]$_0$ died at [90]$_1$" and ``[Chuck Berry]$_0$ lived until [90]$_1$". The predicate paraphrases, i.e. pairs of predicate templates (like ``[a$_0$] died at [a$_1$]" and ``[a$_0$] lived until [a$_1$]") are then aggregated and ranked with the following (unsupervised) heuristic scoring function:

\begin{center}
    $s=n\cdot(1 + \frac{d}{N})$.
\end{center}

This score is proportional to the number of supporting pair instances in which the two templates were paired ($n$), as well as the number of different days in which such pairings were found ($d$), where $N$ is the number of days the resource is collected. The Chirps resource provides the scored predicate paraphrases as well as the supporting pairs for each paraphrase. 

Chirps has acquired more than 5 million distinct paraphrase pairs over the last 3 years. Human evaluation showed that this scoring is effective and that the percentage of correct paraphrases is higher for highly scored paraphrases. At the same time, due to the heuristic collection and scoring of predicate paraphrases in Chirps, entries in the resource may suffer from two types of errors: (1) type 1 error, i.e., the heuristic recognized pairs of non-paraphrastic predicates as paraphrases. This happens when the same arguments participate in multiple different events, as in the following paraphrases: ``[Police]$_0$ arrest [man]$_1$'' and ``[Police]$_0$ shoot [man]$_1$''; and (2) type 2 error, when the scoring function assigned a low score to a rare but correct paraphrase pair, as in ``[a$_0$] outgun [a$_1$]'' and ``[a$_0$] outperform [a$_1$]'', for which only a single supporting pair was found.

\section{Chirps*: Leveraging Coreference Information for Paraphrasing}
\label{sec:improving_chirps}
Our goal in this section is to improve paraphrase scoring, in the context of Chirps, while leveraging available information and methods for event cross-document coreference resolution.
To that end, we introduce Chirps*, a new supervised scorer for Chirps candidate paraphrases, whose novelties are two fold. First, we extract a richer feature representation for a candidate paraphrase pair (Section~\ref{sec:features}), which is fed into a supervised classifier for the candidates. Second, we collect, semi-automatically, distantly supervised training data for paraphrase classification, which is derived from the ECB+ cross-document coreference training set, leveraging the close relationship between the two tasks (Section~\ref{sec:distant_supervision}). Finally, we provide some  implementation details (Section~\ref{sec:model_for_chirps}).

\subsection{Features}
\label{sec:features}
As described above, the original heuristic Chirps scorer relied only on a couple of features to score a candidate paraphrase pair. Our goal is to obtain a richer signal about the likelihood of a candidate predicate pair to indeed be paraphrastic. To that end, we collect a set of features from the available data, with a focus on assessing whether the instances from which the candidate pair was extracted indeed constitute cross-document coreferences. 

Each candidate paraphrase pair consists of two predicate templates $p_1$ and $p_2$, accompanied by the $n$ supporting pair instances for the pair, each consisting a pair of argument terms, associated with this predicate paraphrase pair: $\operatorname{support-pairs}(p_1, p_2) = \{(t_1^1, t_2^1), ..., (t_1^n, t_2^n)\}$. Each tweet included in Chirps links to a news article, whose content we retrieve. When representing a pair of predicate templates, we include both local features (based on a single supporting pair) and global features (based on all supporting pairs). 

\begin{table*}[t]
 \small
 \centering
 \setlength{\tabcolsep}{6pt}
 \setlength{\extrarowheight}{.5em}
 \begin{tabular}{m{5em}m{14em}m{25em}}
 \toprule
& \textbf{Name} & \textbf{Description}  \\
 \toprule
 \multirow{9}{5em}{Chirps-based Features} & \textbf{\# Templates} & The number of different predicate paraphrase pairs with $p_1$ and $p_2$ as predicates, regardless of argument ordering, e.g. ``{[}a$_0${]} release {[}a$_1${]}'' / ``{[}a$_0${]} reveal {[}a$_1${]}'' and ``release {[}a$_0${]} {[}a$_1${]}'' / ``reveal {[}a$_0${]} {[}a$_1${]}'' are both counted. \\
 & \textbf{\# Supporting pairs} & The total number of support pairs of $p_1$ and $p_2$ across the template variants. \\
 & \textbf{\# Days} & The total number of days $d$ that $p_1$ and $p_2$ was matched in Chirps across the template variants. \\
 & \textbf{\# Available supporting pairs} & The number of support pairs of $p_1$ and $p_2$ across the template variants \emph{that were still available to download}. \\
 & \textbf{\# Days of available pairs} & The total number of days $d$ in which the support pairs above occurred in the available tweets. \\
 & \textbf{Score} & The maximal Chirps score across the template variants. \\
 \midrule
 \multirow{2}{5em}{\nameref{ne_feature}} & \textbf{\# NEC above threshold} & Number of pairs with NEC score of at least $T$. \\
 & \textbf{Average above threshold} & Average of NEC scores for pairs with a score of at least $T$. \\
 \midrule
 \multirow{6}{5em}{\nameref{cdcr_featue}} & \textbf{\# Event Perfect} & Number of event pairs with perfect match. \\
 & \textbf{\# Event No Match} & Number of event pairs with no match. \\
 & \textbf{\# Entity Perfect} & Number of entity pairs with perfect match.  \\
 & \textbf{\# Entity Reverse} & Number of entity pairs with reverse match.  \\
 & \textbf{\# Entity No Match} & Number of entity pairs with no match. \\
 & \textbf{\# Perfectly Clustered + NEC} & The number of pairs with NEC score of at least $T$ \textbf{and} perfect clustering for event coreference resolution. \\
 \midrule
 \multirow{2}{5em}{\nameref{cc_feature}} & \textbf{\# Connected components} & The number of connected components in $G_{p_1, p_2}$. \\
 & \textbf{Average component size} & The average size of connected components in $G_{p_1, p_2}$. \\
 \midrule
 \nameref{clique_feature} & \textbf{\# In Clique} & The number of pairs in $\operatorname{support-pairs}(p_1, p_2)$ that are in a clique.  \\
 \bottomrule
 \end{tabular}
 \caption{All 17 features used by our scorer, for a given predicate paraphrase pair $p_1$, $p_2$, as detailed in Section~\ref{sec:features}.}
\label{table:features}
\end{table*}

Table \ref{table:features} presents our 17 features, yielding a feature representation $f_{p_1, p_2} \in \mathcal{R}^{17}$ for a paraphrase pair, grouped by different sources of information. The first group includes features derived from the statistics provided by the original Chirps resource. The other 4 sources of information are described in the following paragraphs.

\paragraph{Named Entity Coverage} \label{ne_feature} While the original Chirps method did not utilize the content of the linked article, we find it useful to retrieve more information about the event. Specifically, it might help mitigating errors in Chirps' argument matching mechanism, which relies on argument alignment considering only the text of the two tweets. We found that the original mechanism worked particularly well for named entities while being more error-prone for common nouns, which might require additional context. 

Given $(t_1^i, t_2^i) \in \operatorname{support-pairs}(p_1, p_2)$, we use SpaCy \cite{spacy2} to extract sets of named entities, $NE_1$ and $NE_2$, from the first paragraph of the news article linked from each tweet,  respectively. We define a Named Entity Coverage score, $\operatorname{NEC}$, as the maximum ratio of named entity coverage of one article by the other:

\vspace{-15pt}
\begin{equation*}
\resizebox{0.94\hsize}{!}{
$
    \operatorname{NEC}(NE_1, NE_2)=\max \left( \frac{ \left |NE_{1}\bigcap NE_{2}  \right |}{\left |NE_{1}  \right |} , \frac{\left |NE_{1}\bigcap NE_{2}  \right |}{\left |NE_{2}  \right |} \right )
$
}
\end{equation*}

We manually annotated a small balanced training set of 121 tweet pairs and used it to tune a score threshold $T=0.26$, such that pairs of tweets whose $\operatorname{NEC}$ is at least $T$ are considered coreferring. Finally, we include the following features: the number of coreferring tweet pairs (whose NEC score exceeds $T$) and the average NEC score of these pairs.

\paragraph{Cross-document Coreference Resolution} \label{cdcr_featue} We apply the state-of-the-art cross-document coreference model from \newcite{barhom-etal-2019-revisiting} to data constructed such that each tweet constitutes a document and each pair of tweets corresponding to $t_1^j$ and $t_2^j$ in 
$\operatorname{support-pairs}(p_1, p_2)$ forms a topic, to be analyzed for coreference. As input for the model, in each tweet, we mark the corresponding predicate span as an event mention and the two argument spans as entity mentions. The model outputs whether the two event mentions corefer (yielding a single event coreference cluster for the two mentions) or not (yielding two singleton clusters). Similarly, it clusters the four arguments to entity coreference clusters.

Differently from Chirps, this model makes its event clustering decision based on the predicate, arguments, and the context of the full tweet, as opposed to considering the arguments alone. Thus, we expect it not to cluster predicates whose arguments match lexically, if their contexts or predicates don't match (first example in Table~\ref{table:chirps_vs_joint_model}). In addition, the model's mentions representation might help to identify lexically-divergent yet semantically-similar arguments (second example in Table~\ref{table:chirps_vs_joint_model}).

\begin{table}[t]
\small
\setlength\tabcolsep{1.5pt}
\begin{tabular}{l}
\toprule
{[}Police]$_0$ [arrest] [two men]$_1$ in incident at Westboro Beach.
\\ 
{[}Police]$_0$ [kill] [man]$_1$ in Vegas hospital who grabbed gun.\\ 
\midrule
{[}Police]$_0$ [arrest] [man]$_1$ in incident at Westboro Beach.\\ 
{[}Officers]$_0$ [seize] [guy]$_1$ in incident at Westboro Beach.
\\ 
\bottomrule
\end{tabular}

\caption{Examples of coreference errors made by Chirps and corrected by \newcite{barhom-etal-2019-revisiting}: 1) false positive: wrong man / two men alignment (disregarding location modifiers). 2) (hypothetical) false negative: lexically-divergent yet semantically-similar arguments.}
\label{table:chirps_vs_joint_model}
\end{table}

For a given pair of tweets, we extract the following binary features with respect to the predicate mentions: \emph{Event Perfect} when the predicates are assigned to the same cluster, and \emph{Event No Match} when each predicate forms a singleton cluster. For argument mentions, we extract the following features: \emph{Entity Perfect} if the two a$_0$ arguments belong to one cluster and the two a$_1$ arguments belong to another cluster; \emph{Entity Reverse} if at least one of the a$_0$ arguments is clustered as coreferring with the a$_1$ argument in the other tweet; and \emph{Entity No Match} otherwise. Also, we extract \emph{Perfectly Clustered with NE Coverage} that combines both Named Entity coverage and coreference-resolution, which count the number of pairs their events are perfectly clustered and with NEC score of at least T.

\paragraph{Connected Components} \label{cc_feature}The original Chirps score of a predicate paraphrase pair is proportional to two parameters: (1) the number of supporting pairs; (2) the ratio of number of days in which supporting pairs were matched relative to the entire collection period. The latter lowers the score of paraphrase pairs which might have been mistakenly aligned on relatively few days (e.g. due to misleading argument alignments in particular events). The number of days in which the predicates were aligned is taken as a proxy for the number of different events in which the predicates co-refer. Here, we aim to get a more reliable partition of tweets to different events by constructing a graph of tweets as nodes, with supporting tweet pairs as edges, and looking for connected components. 

To that end, we define a bipartite graph ${G_{p_1, p_2}=(V, E)}$ for a candidate paraphrase pair, where ${V = \operatorname{tweets}(p_1, p_2)}$ contains all the tweets in which $p_1$ or $p_2$ appeared, and ${E = \operatorname{support-pairs}(p_1, p_2)}$. We compute $C$, the number of connected components in $G_{p_1, p_2}$, and define the following group: $ConComp = \{c \in C : |c| > 2\}$, which represents the number of connected components with size greater than 2. From this group we derive two features ${\operatorname{\#connected}(p_1, p_2)= |ConComp|}$ which represents the number of the connected components and  ${\operatorname{avg\_connected}(p_1, p_2)}$,
which is the average size of the connected components in the graph. A larger number of connected components indicates that the two predicates were aligned across a large number of likely different events. 

\paragraph{Clique} \label{clique_feature}We similarly build a global tweet graph for all the predicate pairs, ${G_{all} = (V', E')}$, where ${V' = \cup_{(p_1, p_2)} \operatorname{tweets}(p_1, p_2)}$, and ${E' = \cup_{(p_1, p_2)} \operatorname{support-pairs}(p_1, p_2)}$. We compute $Q$, the set of cliques in ${G_{all}}$ of size greater than 2. We assume that a pair of tweets are more likely to be coreferring if they are part of a bigger clique, whereas if they were extracted by mistake they wouldn't share many neighbors. We extract the following feature of clique coverage for a candidate paraphrase pair:  $\operatorname{CLC}(p_1, p_2)= |\{t_1^j, t_2^j \in  \operatorname{support-pairs}(p_1, p_2) : \exists q \in Q \text{~such that~} t_1^j \in q \wedge t_2^j \in q \}|$.

\subsection{Distantly Supervised Labels} 
\label{sec:distant_supervision}

In order to learn to score the paraphrases, we need gold standard labels, i.e., labels indicating whether a pair of predicate templates collected by Chirps is indeed a paraphrase. Instead of collecting manual annotations for a sample of the Chirps data, we chose a low-budget distant supervision approach. To that end, we leverage the similarity between the predicate paraphrase extraction and the event coreference resolution tasks, and use the annotations from the ECB+ dataset.

Our dataset consists of the predicate paraphrases from Chirps that appear in ECB+ (denoted ch-ECB+). As positive examples we consider all pairs of predicates $p_1, p_2$ from Chirps that appear in the same event cluster in ECB+, e.g., from \{\textit{talk, say, tell, accord to, confirm}\} we extract (\textit{talk, say}), (\textit{talk, tell}), ..., (\textit{accord to, confirm}). 
\begin{table}[t]
\centering
\small
\begin{tabular}{@{}llll@{}}
\toprule
      &             & Pre Annotation & Post Annotation \\ \midrule
\textbf{Train} & \# positive & 266            & 803             \\
      & \# negative & 2040           & 1056            \\\hline
\textbf{Dev}   & \# positive & 93             & 222             \\
      & \# negative & 539            & 318             \\\hline
\textbf{Test}  & \# positive & 131            & 352             \\
      & \# negative & 758            & 411             \\ \bottomrule
\end{tabular}
\caption{Statistics of the paraphrase scorer dataset. The difference in size before and after the annotation is due to omitting examples with less than 3 supporting pairs.}
\label{table:annotation_statistics}
\end{table}

Obtaining negative examples is a bit trickier. We consider as negative example candidates pairs of predicates $p_1, p_2$ from Chirps, which are \emph{under the same topic}, but in different event clusters in ECB+, e.g., given the clusters \{\textit{specify, reveal, say}\}, and \{\textit{get}\}, we extract (\textit{specify, get}), (\textit{reveal, get}), and (\textit{say, get}). 

Note that the ECB+ annotations are context-dependent. Thus a pair of predicates that are in principle coreferable may be annotated as non-coreferring in a given context. To reduce the rate of such false-negative examples, we validated all the candidate negative examples and a sample of the positive examples using Amazon Mechanical Turk. Following \newcite{shwartz-etal-2017-acquiring}, we annotated the templates while presenting 3 argument instantiations from their original tweets. Thus, we only included in the final data predicate pairs with at least 3 supporting pairs. We required that workers have 99\% approval rate on at least 1,000 prior tasks and pass a qualification test. 

\setlength\tabcolsep{2pt}
\begin{table}[t!]
\centering
\small
\begin{tabular}{@{}l|ll|c}
\toprule
              & \multicolumn{2}{c}{ch-ECB+ Test Set} & ch-Random \\
              Model& AP      & Accuracy (P / R / $F1$)  &  AP \\
              \midrule
GloVe      & 50.7 & 55.7   (58.1 / 14.2 / 22.8) & 50.1\\
Chirps     & 62.5 & 60.4   (59.9 / 45.7 / 51.6) & 51.4\\
Chirps* & \textbf{80.0} &\textbf{73.8}   (\textbf{74.1} / \textbf{66.5} / \textbf{70.1}) & \textbf{59.5}\\ \bottomrule
\end{tabular}
\caption{Ranker (AP) and classification (Accuracy, Precision, Recall, $F1$) results on ch-ECB+ test set (middle column),  and Ranker results on ch-Random, a subset of 500 randomly selected predicate pairs from the entire Chirps resource (right column). }
\label{table:classifier_evaluation}
\end{table}

Each example was annotated by 3 workers. We aggregated the per-instantiation annotations using majority vote and considered a pair as positive if at least one instantiation was judged as positive. The data statistics are given in Table~\ref{table:annotation_statistics}. This validation phase balanced the positive-negative proportion of instances in the data, from approximately 1:7 to approximately 4:5.

\subsection{Model}
\label{sec:model_for_chirps}

We trained a random forest classifier \cite{breiman2001random} implemented by the scikit-learn framework \cite{scikit-learn}. To tune the hyper-parameters, we ran a 3 fold cross-validation randomized search, yielding the following values: 157 estimators, max depth of 8, minimum samples leaf of 1, and min samples split of 10.\footnote{We chose random forest over a neural model because of the small size of the training set.} 

\begin{table}[t!]
\centering
\scriptsize
\begin{tabular}{l|lll}
\toprule
 Rank & \textbf{GloVe} & \textbf{Chirps} & \textbf{Chirps*}  \\
 \midrule
1 & \cellcolor{blue!10} take over / take up & \cellcolor{blue!10} announce / unveil &
\cellcolor{blue!10} launch / unveil \\
11 & begin / continue & \cellcolor{blue!10}  introduce / unveil & \cellcolor{blue!10}  buy / purchase \\
21 & find / go & accuse / warn & \cellcolor{blue!10} add / introduce \\
31 & move / plan & \cellcolor{blue!10} launch / unveil & \cellcolor{blue!10} announce / launch \\
41 & buy / pay & \cellcolor{blue!10} announce / enter & accuse / warn \\
51 & \cellcolor{blue!10} die / kill & accuse / threaten & \cellcolor{blue!10} cast / set \\
61 & \cellcolor{blue!10} announce / confirm & arrest / charge & say / warn \\
71 & deal / say & kill / tell & \cellcolor{blue!10} fire / use \\
81 & \cellcolor{blue!10} condemn / deny & \cellcolor{blue!10} attack / strike & \cellcolor{blue!10} cause / trigger \\
91 & believe / move & \cellcolor{blue!10} claim / kill & \cellcolor{blue!10}  refresh / upgrade \\
\bottomrule
\end{tabular}
\caption{Highly ranked paraphrases from the intersection of the candidate coreference pairs in the ECB+ test set and Chirps (ch-ECB+ test set). The table presents 10 highly ranked paraphrases, as ranked by each of GloVe, Chirps and our method, taken from ranks 1, 11, 21 and so on in each ranking. Pairs labeled as positive in our gold dataset are highlighted in purple.}
\label{table:ranker_results}
\end{table}

\setlength\tabcolsep{2pt}
\begin{table}[t]
\centering
\small
\begin{tabular}{llllll}
\toprule
\textbf{Ablated Feature Set} & \textbf{AP} & \multicolumn{4}{l}{\textbf{Classification Metrics}} \\ 
& & \textbf{Acc.} & \textbf{P} & \textbf{R} & $\mathbf{F_1}$ \\
\midrule
Chirps & 73.94 & 72.96 & 74.36 & 52.25 & 61.38 \\
Named Entity Coverage      & 76.43 & \textbf{74.44} & 73.86 & \textbf{58.56} & \textbf{65.33} \\
Coreference & 75.22 & 72.04 & 72.32 & 51.8 & 60.37 \\
Connected Component & 76.59 & 73.51 & \textbf{74.53} & 54.05 & 62.66 \\
Clique & 76.78 & 73.19 & 72.51 & 55.85 & 63.1 \\
\midrule
All Features & \textbf{77.13} & 73.88 & 73.96 & 56.3 & 63.94 \\ \bottomrule
\end{tabular}
\caption{Ablation test results on the ch-ECB+ dev set for the ranking and classification models.}
\label{table:ablation_test}
\end{table}
\subsection{Evaluation}
\label{sec:intrinsic_eval}

We used the model for two purposes: (1) classification: determining if a pair of predicate templates are paraphrases or not; and (2) ranking the pairs based on the predicted  positive class score. We consider the ranking evaluation as more informative, as we expect the ranking to reflect the number of contexts in which a pair of predicates may be coreferring. That is, predicate pairs that are coreferring in many contexts will be ranked higher than those that are coreferring in just a few contexts. 

We compare our model with two baselines: the original Chirps scores, and a baseline that assigns each pair of predicates the cosine similarity scores between the predicates using GloVe embeddings \cite{pennington-etal-2014-glove}.\footnote{We were motivated to compare with Glove since this resource is utilized as a lexical representation in the state-of-the-art system of \cite{barhom-etal-2019-revisiting}, which is utilized in the next section. Here, multi-word predicates were represented by the average of their Glove word vectors.} For the classification decisions made by the two baseline scores (Chirps score and cosine similarity for Glove vectors), we learn a threshold that yields the best accuracy score over the train set, above which a pair of predicates is classified as positive.

\begin{table*}[t]
    \small
    \centering
    \begin{tabular}{m{20em}|ccc|ccc|ccc|c}
        \toprule
         & & MUC & & & B$^3$ & & & CEAF-$e$ & &  CoNLL \\
        \textbf{Model} & R & P & $F_1$ & R & P & $F_1$ & R & P & $F_1$ & $F_1$\\
        \hline
        \textsc{Lemma Baseline} &76.5&79.9&78.1&71.7&85&77.8&75.5&71.7&73.6&76.5\\
        \newcite{barhom-etal-2019-revisiting} &77.6&84.5&80.9&\bf 76.1&85.1& 80.3&81&73.8&77.3&79.5\\
        \newcite{barhom-etal-2019-revisiting}\textsc{ + Chirps* Features} &\textbf{78.7}&\textbf{84.67}&\textbf{81.61}&75.87&\textbf{85.91}&\textbf{80.5}&\textbf{81.09}&\textbf{74.77}&\textbf{77.8}&\textbf{80.0}\\
        \bottomrule
        
    \end{tabular}
    \vspace*{-7pt}
    \caption{Event mentions coreference-resolution results on ECB+ test set.}
    \label{table:model_results}
    \vspace*{-7pt}
\end{table*}

Table~\ref{table:classifier_evaluation} displays the accuracy, precision, recall and $F_1$ scores for classification evaluation and the Average Precision (AP) for ranking evaluation. Our scorer dramatically improves upon the baselines in all metrics. 

To show that the improved scoring generalizes beyond examples that appear in the ECB+ dataset, we selected a random subset of 500 predicate pairs with at least 6 support pairs from the entire Chirps resource and annotated them in the same method described in Section~\ref{sec:distant_supervision}. The ranker evaluated on this subset gained 8 points in AP, relative to the original Chirps ranking. All results are statistically significant using bootstrap and permutation tests with $p < 0.001$ \cite{dror-etal-2018-hitchhikers}. 

Table~\ref{table:ranker_results} exemplifies highly ranked predicate pairs by our Chirps* scorer, the original Chirps scorer and the GloVe scorer, which illustrates the improved ranking performance of Chirps* (as measured in table \ref{table:classifier_evaluation} by the AP score).

\paragraph{Ablation Test} To evaluate the importance of each type of feature, we perform an ablation test. Table~\ref{table:ablation_test} displays the performance of various ablated models, each of which with one set of features (Section~\ref{sec:features}) removed from the representation. In the classification task, removing the named entity coverage features somewhat improved the performance, mostly by increasing the recall. However, in terms of the (primary) ranking evaluation, each set of features contributed to the performance, with the full model performing best.

\section{Leveraging a Paraphrasing Resource to Improve Coreference}
\label{sec:using_chirps}
In Section~\ref{sec:improving_chirps} we showed that leveraging CD event coreference annotations and model improves predicate paraphrase ranking. In this section, we show that this co-dependence can be used in both directions, and that using Chirps* as an external resource can improve the performance of a CD model. 

As a preliminary analysis, we computed Chirps' coverage of lexically-divergent pairs of co-referring event mentions in ECB+. We found approximately 30\% coverage overall and above 50\% coverage for coreferring verbal mentions.
\footnote{Non-verbal mentions in ECB+ include nominalizations (\textit{investigation}), names (\textit{Oscars}) acronyms (\textit{DUI}), and more. Chirps, by design, consists of verbal predicates only.}
This indicates a substantial coverage of the lexically-divergent positive coreferrability decisions that need to be made in ECB+. In absolute numbers, Chirps covers about 370 lexically-divergent pairs of coreferring event mentions appearing in the ECB+ training set, and about 200 in the test set.

\subsection{Integration Method}
\label{sec:integration_method}

The state-of-the-art CD coreference resolution model, by \newcite{barhom-etal-2019-revisiting}, trained a pairwise mention scoring function, $MLP_{scorer}(m_i, m_j)$, which predicts the probability that two mentions $m_i$, $m_j$ refer to the same event. The mention representation includes a lexical component (GloVe embeddings) as well as a contextual component \cite[ELMo embeddings,][]{peters2018deep}. The mention pair representation $\vec{v}_{i, j}$, which is fed to the pairwise scorer, combines the two separate mention representations. 

\begin{figure}[!t]
    \hspace{-5pt}
    \includegraphics[width=.42\textwidth]{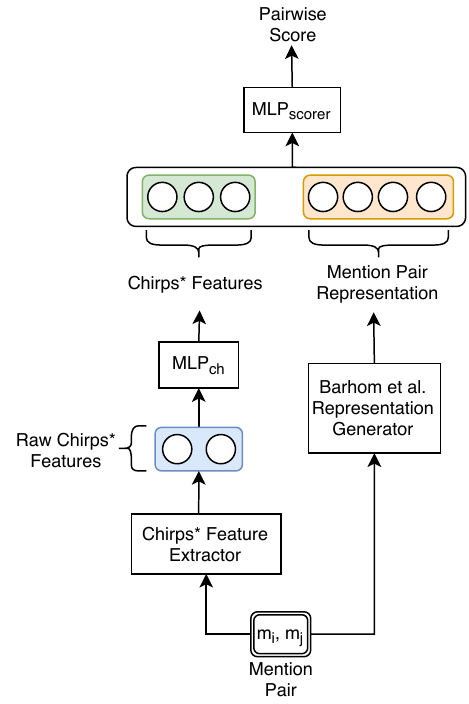}
    \caption{An illustration of the integrated mention pair scorer. The right vector is the original mention pair vector from \citet{barhom-etal-2019-revisiting}, and the left one is our Chirps* extension, which is transformed through $MLP_{ch}$ into the same embedding space. The two vectors are concatenated to form the mention pair representation, which is fed to the scoring function  $MLP_{scorer}$.}
    \label{fig:modelFigure}
\end{figure}

We extended the model by changing the input to the pairwise event mention scoring function to include information regarding the mention pair from Chirps*, as illustrated in Figure~\ref{fig:modelFigure}. We defined $\vec{{v'}_{i, j}} = [\vec{v}_{i, j};\vec{c}_{i, j}]$, where $\vec{c}_{i, j}$ denotes the Chirps* features, computed in the following way: 

\begin{equation*}
    \vec{c}_{i, j} = \begin{cases}
    MLP_{ch}(\vec{f}_{m_i, m_j}) & \text{if~} m_i, m_j \in \text{Chirps}\\
    MLP_{ch}(\vec{0})              & \text{otherwise}
    \end{cases}
\end{equation*}

\noindent $\vec{f}_{m_i, m_j} \in \mathcal{R}^{17}$ is the feature vector representing a pair of predicates $(m_i, m_j)$ for which there is an entry in Chirps, otherwise the input is a zero vector. $MLP_{ch}$ is an MLP with a single hidden layer of size 50 and output layer of size 100, which is used to transform the discrete values in $\vec{f}_{m_i, m_j}$ into the same embedding space of $\vec{v}_{i, j}$. The rest of the model remains the same, including the model architecture, training, and inference.\footnote{We also tried to incorporate only the final Chirps* score into the mention pair representation, but the performance improvement was smaller.}

\subsection{Evaluation}
\label{sec:extrinsic_eval}

We evaluate the event coreference performance on ECB+ using the official CoNLL scorer \cite{pradhan-etal-2014-scoring}. The reported metrics are \textbf{MUC} \cite{vilain1995model}, $\mathbf{B^3}$ \cite{bagga1998algorithms}, \textbf{CEAF-$e$} \cite{luo-2005-coreference} and \textbf{CoNLL $F_1$} (the average of MUC, $B^3$ and CEAF-$e$ scores).

We compare the integrated model to the original model and to the lemma baseline which clusters together mentions that share the same mention-head lemma. The results in Table~\ref{table:model_results} show that the Chirps-enhanced model provides an improvement of 3.5 points over the lemma baseline and a small improvement upon \newcite{barhom-etal-2019-revisiting} in all $F_1$ score measures. The greatest improvement is in the link-based MUC measure, which counts the number of corresponding links between the mentions. The Chirps component helps link more coreferring mentions (improving recall) and prevents the linking of some wrong mentions (improving precision). 

Although the gap between our model and the original model by \newcite{barhom-etal-2019-revisiting} is statistically significant (bootstrap and permutation tests with $p < 0.001$), it is rather small. We can attribute it partly to the coverage of Chirps over ECB+ (around 30\%), which entails that the majority of event mention pairs still have the same representation as in the original model. We also note that ECB+ suffers from annotation errors, as was observed by \newcite{barhom-etal-2019-revisiting} and others.

\section{Conclusion and Future Work}
\label{sec:conclusion}
We studied the synergy between the tasks of identifying predicate paraphrases and event coreference resolution, both concerned with matching the meanings of lexically-divergent predicates, and showed that they can benefit each other. Using event coreference annotations as distant supervision, we learned to re-rank predicate paraphrases that were initially ranked heuristically, and managed to increase their average precision substantially. In the other direction, we incorporated knowledge from our re-ranked predicate paraphrases resource into a model for event coreference resolution, yielding a small improvement upon previous state-of-the-art results. We hope that our study will encourage future research to make further progress on both tasks jointly.

\section*{Acknowledgements}
\label{sec:acks}
This work was supported in part by grants from Intel Labs, Facebook, the Israel Science Foundation grant 1951/17, the Israeli Ministry of Science and Technology and the German Research Foundation through the German-Israeli Project Cooperation (DIP, grant DA 1600/1-1).

\bibliography{main}
\bibliographystyle{acl_natbib}

\end{document}